\documentclass[letterpaper, 10pt, conference]{ieeeconf}

\IEEEoverridecommandlockouts   %
\usepackage{amsmath}
\usepackage{amssymb}
\usepackage{amsfonts}
\usepackage{booktabs}
\usepackage{tabularx}
\usepackage{multirow}
\usepackage{graphicx}
\usepackage{subcaption}
\usepackage{algorithm}
\usepackage{algpseudocode}
\usepackage{overpic}
\usepackage[bookmarks=true,colorlinks,allcolors=black]{hyperref}
\usepackage[table]{xcolor}
\usepackage[utf8]{inputenc}
\usepackage[T1]{fontenc}
\usepackage[english]{babel}
\usepackage{tikz}
\usetikzlibrary{arrows.meta,calc,decorations.pathreplacing}
\usepackage{pgfplots}
\pgfplotsset{compat=1.18}
\usepackage{etoolbox}

\makeatletter
\let\NAT@parse\undefined
\makeatother
\usepackage[numbers,sort&compress]{natbib}

\DeclareCaptionLabelSeparator{ieee}{.~~}
\newcommand{\algcommentstyle}[1]{{\color{black!60}\itshape//~#1}}
\algrenewcommand{\algorithmiccomment}[1]{\hfill\algcommentstyle{#1}}
\makeatletter
\newcommand{\LineComment}[1]{\Statex \hskip\ALG@thistlm \algcommentstyle{#1}}
\g@addto@macro\fs@ruled{\def\@fs@pre{\kern2mm\hrule height.8pt depth0pt \kern2pt}}
\makeatother

\newcommand{\R}{\mathbb{R}}
\newcommand{\E}{\mathbb{E}}
\newcommand{\M}{\mathcal{M}}
\newcommand{\SO}[1]{\mathrm{SO}(#1)}
\newcommand{\SE}[1]{\mathrm{SE}(#1)}

\newcommand{\Exp}{\mathrm{Exp}}
\newcommand{\Log}{\mathrm{Log}}

\newcommand{\sg}{\mathrm{sg}}
\newcommand{\Sph}[1]{\mathcal{S}^{#1}}

\definecolor{purduegold}{HTML}{C28E0E}
\hypersetup{
  bookmarksopen,
  bookmarksnumbered,
  colorlinks=true,
}

\definecolor{modecol}{HTML}{1F3864}
\definecolor{eulercol}{HTML}{C1121F}
\definecolor{flowcol}{HTML}{2A7F7F}
\definecolor{mfdline}{HTML}{BFBFBF}

\definecolor{hicell}{HTML}{ebd99f}

\newcommand{\cmidrowshift}{-.5\dimexpr\aboverulesep+\cmidrulewidth+\belowrulesep\relax}
\newcommand{\lasatrim}{40pt}
\newcommand{\lasafigaux}[2]{\includegraphics[width=\linewidth,trim=#1 #1 #1 #1,clip]{#2}}
\newcommand{\lasafig}[1]{\expandafter\lasafigaux\expandafter{\lasatrim}{#1}}

\title{\LARGE \bf
Faster Visuomotor Policy Learning on Action Manifolds\\
via Riemannian MeanFlow
}

\author{S. Talha Bukhari, Austin Garrett, Yi Wei, Ruiqi Ni, Zachary Kingston, and Aniket Bera%
\thanks{Authors are with the Department of Computer Science, Purdue University, West Lafayette, IN 47907, USA.
\texttt{\small \{bukhars, garre155, wei577, ni117, zkingston, aniketbera\}@purdue.edu}}
}

\begin{document}

\maketitle
\thispagestyle{empty}
\pagestyle{empty}

\begin{abstract}
Visuomotor policies learn a direct map from raw sensory observations to robot action sequences.
Policies based on Diffusion and Flow Matching capture the multimodal distribution over action sequences in an end-to-end manner.
This expressivity comes at the cost of multi-step numerical integration of the learned vector field for action generation, which can be expensive and time-consuming, impeding fast control rates required in robotics applications.
Furthermore, robot action sequences are usually defined on a smooth, differentiable manifold, requiring that the learned policy respects the intrinsic geometry of the robot's action space.
Here, we present Riemannian MeanFlow Policy (RMFP), which learns the conditioned flow map of the probability path on the robot action manifold.
Our formulation employs a flow map consistency objective grounded in the data by a Riemannian Conditional Flow Matching anchor.
The flow map consistency condition is stable to train and constrains the learned model to finite-time transport, which yields on-manifold action sequence generation with as few as one network function evaluation.
We present results on the spherical LASA and Push-T benchmarks, on the Tool Hang and Transport tasks of the Robomimic suite, and on the Franka Kitchen task with manifold-constrained action generation, and demonstrate that RMFP attains performance competitive with prior work at a lower sampling cost.
We also employ RMFP on a real-world robotic manipulation task to demonstrate fast action generation under imperfect sensor measurements in the physical world.

\end{abstract}

\section{Introduction}\label{sec:introduction}

Visuomotor policy learning infers a distribution over robot action sequences from raw sensor observations, and the demonstrations that supervise it are multimodal, since several distinct action sequences accomplish the same task from the same observation.
Generative policies model that distribution directly and avoid the mode averaging incurred by a regression policy trained under a unimodal likelihood~\citep{Chi2025DiffusionPolicy}.
The actions are not Euclidean in general: an end-effector command specifies a position, an orientation and a gripper state, and the action space is therefore a product manifold such as $\R^3 \times \Sph{3} \times \R^2$~\citep{Ding2025RFMP}.
A policy that treats those coordinates as $\R^9$ assigns probability mass off the manifold and requires a projection at execution that its training objective does not model.

Diffusion~\citep{Ho2020DDPM, Song2021Score} and Flow Matching~\citep{Lipman2023FlowMatching} provide the generative backbone of most current manipulation policies~\citep{Chi2025DiffusionPolicy, Ze2024DP3, Black2025Pi0}.
Both generate by integrating a learned field over a time interval, which yields broad mode coverage and a stable regression objective at a cost of one network function evaluation (NFE) per integration step.
Riemannian Flow Matching Policy (RFMP) transfers this construction to the action manifold by defining the conditional paths as geodesics and integrating the learned velocity field with an exponential-map Euler scheme~\citep{Braun2024RFMP, Ding2025RFMP}.
While prior work shows that Flow Matching tolerates a coarse grid far better than diffusion does~\citep{Ding2025RFMP}, the budget is nonetheless a property of the solver rather than of the objective, and the reported scores are not monotone in it; it must therefore be tuned per task.
Sampling with a coarse grid on a Riemannian manifold can lead to mode collapse, which a task success rate alone does not reveal.

\begin{figure}[t]
  \centering
  \begin{subfigure}[b]{0.49\linewidth}
    \centering
    \lasafig{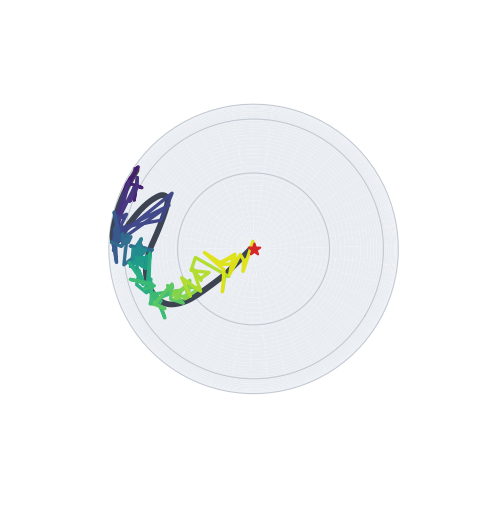}
    \caption{RFMP}\label{fig:lasa-rfm}
  \end{subfigure}
  \hfill
  \begin{subfigure}[b]{0.49\linewidth}
    \centering
    \lasafig{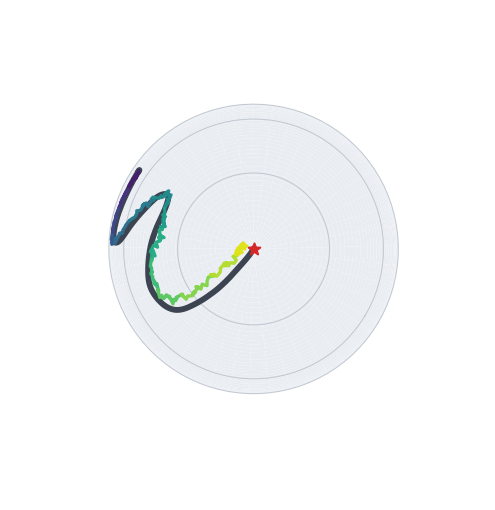}
    \caption{RMFP (Ours)}\label{fig:lasa-rmf}
  \end{subfigure}
  \caption{We present Riemannian MeanFlow Policy (RMFP) for fast action sequence generation on manifolds.
  Here we show 1-NFE trajectory generation for the LASA shape \textsf{W} projected onto $\Sph{2}$~\citep{Braun2024RFMP}.
  Demonstrations from the dataset are shown in \emph{black} and the executed trajectory is colorized by time-parameterization.
  \textcolor{red}{$\star$} denotes the north pole.
  As shown, RMFP traces a smoother path while abiding by the demonstrations.
  Median DTW is $0.023$ for RMFP against $0.079$ for RFMP.}
  \label{fig:lasa}
  \vspace{-1mm}
\end{figure}

Few-step generative models instead learn the flow map, the transport that the velocity field induces over a finite interval, so that one NFE performs a finite-time jump.
Consistency models enforce self-consistency along the trajectories of a probability-flow ODE~\citep{Song2023Consistency}, and their policy instantiation distils a pretrained Diffusion Policy~\citep{Prasad2024ConsistencyPolicy}, which adds a sequential training stage and bounds the student by the teacher.
Shortcut models~\citep{Frans2025Shortcut} and MeanFlow~\citep{Geng2025MeanFlow} remove the teacher, and MeanFlow regresses the average velocity over an interval, a property of the probability path whose optimum is characterized independently of the network.
Three concurrent formulations carry flow map learning to Riemannian manifolds~\citep{Davis2026GFM, Woo2026RiemannianMeanFlow, Zhong2026RMFManifolds}, and all three report on geospatial, biomolecular or synthetic manifold data.
The transfer to policy learning remains open: MeanFlow policies are formulated for Euclidean action spaces~\citep{Sheng2026MP1}, and neither the objective nor its parameterizations have been evaluated on the action sequences and product manifolds that a manipulation policy generates.

In this work, we present Riemannian MeanFlow Policy (RMFP), which learns the conditioned flow map of the generative probability path on the robot action manifold rather than the instantaneous velocity field that induces it.
The formulation is obtained by coupling a flow map consistency term with a Riemannian flow matching anchor, allowing few-step generation at a performance competitive with multi-step state-of-the-art methods for on-manifold action generation.
Training needs no teacher and no differential terms, and therefore remains single-stage and first-order, at a cost of four NFEs per gradient step against the one that RFMP pays.
We instantiate the policy network under velocity-prediction and endpoint-prediction formulations, which impose the manifold constraint at different points of the forward pass, and evaluate both across tasks.

We present the following contributions:
\begin{itemize}
  \item A flow map formulation of visuomotor policy learning on a product action manifold, coupling the algebraic semigroup identity with a Riemannian Conditional Flow Matching anchor.
  This enables single-stage policy training without a teacher and without differentiating the geometric maps.
  \item Endpoint and velocity parameterizations of the network head on the action manifold, an analysis of what each regresses away from the data endpoint, and a comparison of their empirical performance.
  \item Extensive evaluation of fast action generation performance of the proposed formulation against prior work on Riemannian policy learning on benchmarks spanning Euclidean, spherical, and product action manifolds, demonstrating its efficacy for robot policy learning across a variety of tasks.
\end{itemize}

\section{Related Work}\label{sec:related_work}

\subsection{Generative Models on Riemannian Manifolds}\label{ssec:related_manifold}

Recent work reformulates generative models on Riemannian manifolds so that the learned distribution is supported on the manifold by construction.
Riemannian continuous normalizing flows parameterize the density as the solution of an ODE constrained to the manifold~\citep{Mathieu2020RCNF}, score-based and variational diffusion models replace the Euclidean forward process with one driven by the manifold's heat kernel~\citep{DeBortoli2022RiemannianScore, Huang2022RiemannianDiffusion}, and Riemannian Flow Matching defines the conditional paths as geodesics and regresses their velocity, which removes the simulation and divergence estimates that the earlier formulations require~\citep{Chen2024RiemannianFM}.
These constructions support generative models on $\SE{3}$ and its factors, for protein backbones~\citep{Yim2023SE3Diffusion, Bose2024FoldFlow} and for robot grasp poses~\citep{Urain2023SE3DiffusionFields}.

The aforementioned methods sample by integrating the learned vector field along the manifold, which costs multiple NFEs per sample.
Three concurrent formulations remove that dependence by learning the flow map.
\citet{Davis2026GFM} generalize the Euclidean flow map framework to arbitrary manifolds with Lagrangian, Eulerian and progressive self-distillation objectives; under specific design choices, the Eulerian objective generalizes consistency training and MeanFlow, and the progressive objective generalizes shortcut models.
\citet{Zhong2026RMFManifolds} define the average velocity by parallel transport and train it with the Riemannian form of the MeanFlow differential identity, which is evaluated by Jacobian--vector products; its two loss terms do not align during learning, and it therefore requires conflict-aware multi-task learning.
\citet{Woo2026RiemannianMeanFlow} derive Eulerian, Lagrangian and semigroup characterizations of the chord-form average velocity, and find that endpoint prediction scales to high dimensions and that the semigroup form with endpoint prediction trains most stably for protein backbone generation.
All three are evaluated on geospatial, biomolecular or synthetic manifold data.
In robotics, concurrent work applies MeanFlow to 6-DoF grasp synthesis on $\SE{3}$~\citep{Kwon2026GraspMeanFlow} and $\SO{3} \times \R^3$~\citep{Bukhari2026LieMeanFlow}.

We adopt the chord-form average velocity and the algebraic semigroup characterization, and apply them to end-to-end robot policy learning on different action manifolds, where the generated object is an observation-conditioned manifold-valued action sequence executed in closed loop.

\subsection{Generative Policy Learning for Robot Manipulation}\label{ssec:related_policy}

Generative visuomotor policies learn a distribution over action sequences from demonstrations and sample from it at execution.
Diffusion Policy~\citep{Chi2025DiffusionPolicy} generates action chunks under receding-horizon control with a denoising backbone, which later work extended with 3D observation encoders~\citep{Ze2024DP3} and scaled to vision-language-action models built on a flow matching action expert~\citep{Black2025Pi0}.
The sampling cost of these policies has been reduced along three lines, all formulated on Euclidean action spaces: adaptive solvers spend integration steps where the conditional variance of the flow is high~\citep{Hu2024AdaFlow}, distillation transfers a pretrained diffusion policy into a consistency student~\citep{Prasad2024ConsistencyPolicy}, and flow map objectives train a few-step generator directly~\citep{Sheng2026MP1}.

On a Euclidean action space, an orientation is produced as unconstrained coordinates and mapped onto the rotation group at execution.
Riemannian Flow Matching Policy places the probability path on the action manifold, so that samples lie on it by construction~\citep{Braun2024RFMP, Ding2025RFMP}; its sampling, however, integrates the learned field over multiple NFEs.
This work learns the flow map on the action manifold under geodesic conditional paths; a one-step sample therefore queries the map that the objective regresses, rather than taking a single exponential-map Euler step along the instantaneous field.
Training is single-stage and teacher-free, and the objective contains no differential terms.

\section{Methodology}\label{sec:method}

\subsection{Background}\label{ssec:background}

\subsubsection{Riemannian Geometry}\label{sssec:riemannian_geometry}
The action space of a manipulator is a smooth manifold rather than a vector space.
Let $\mathcal{A}$ denote the manifold of a single action, which for an end-effector controller is the product $\mathcal{A} = \R^3 \times \Sph{3} \times \R^2$ of a position, a unit quaternion and two finger positions~\citep{Ding2025RFMP}.
A policy with prediction horizon $T_p$ emits an \emph{action chunk}, and the object it generates is therefore one point of the product manifold $\M = \mathcal{A}^{T_p}$, which carries the product metric $g$ induced by its factors.
We write $T_a\M$ for the tangent space at $a \in \M$, and $\langle \cdot, \cdot \rangle_a$ and $\|\cdot\|_a$ for the inner product and norm that $g$ induces there.
The exponential map $\Exp_a : T_a\M \to \M$ carries a tangent vector to the endpoint of the geodesic that leaves $a$ with that initial velocity, and the logarithmic map $\Log_a : \M \to T_a\M$ inverts it within the injectivity radius.
Both are available in closed form on the factors used here.
The geodesic from $a_0$ to $a_1$ is given as $\gamma(\tau) = \Exp_{a_0}\!\big(\tau \Log_{a_0}(a_1)\big)$ for $\tau \in [0,1]$, and every construction below uses it as the conditional path.

\subsubsection{Riemannian Flow Matching Policy}\label{sssec:rfmp}
Riemannian Flow Matching Policy (RFMP) learns a time-dependent vector field that transports a tractable prior on $\M$ to the demonstration distribution~\citep{Braun2024RFMP, Ding2025RFMP}.
Let $\rho_0$ be the prior, a standard Gaussian on each Euclidean factor and a wrapped Gaussian on $\Sph{3}$, and let $\rho_1(\cdot \mid o)$ be the distribution of demonstrated action chunks under observation $o$.
Drawing $a_0 \sim \rho_0$ and $a_1 \sim \rho_1(\cdot \mid o)$ independently, and a time $s \sim \mathcal{U}[0,1]$, the conditional path is the geodesic $a_s = \Exp_{a_0}\!\big(s \Log_{a_0}(a_1)\big)$, whose velocity $\dot a_s \in T_{a_s}\M$ is the regression target of
\begin{equation}
  \mathcal{L}_{\mathrm{RCFM}}(\theta) = \E \left[ \big\| v_\theta(a_s, s \mid o) - \dot a_s \big\|_{a_s}^2 \right].
  \label{eq:rcfm}
\end{equation}
The minimizer of~\eqref{eq:rcfm} is the marginal velocity field 
\begin{equation}
    v(a_s, s \mid o) = \E[\dot a_s \mid a_s, o],
\end{equation}
which is the quantity the sampler integrates.
Generation partitions $[0,1]$ by a grid $0 = t_0 < \cdots < t_T = 1$ and advances
\begin{equation}
  a_{t_{k+1}} = \Exp_{a_{t_k}}\!\big((t_{k+1} - t_k)\, v_\theta(a_{t_k}, t_k \mid o)\big)
  \label{eq:rfmp_sampling}
\end{equation}
from $a_{t_0} \sim \rho_0$.
This costs $T$ NFEs, and each step is a first-order approximation of the transport over its interval.

\subsection{Riemannian MeanFlow Policy}\label{ssec:rmfp}

\subsubsection{MeanFlow Objective}\label{sssec:meanflow}
\begin{figure}[!t]
\centering\vspace{2mm}
\begin{tikzpicture}[>={Stealth[length=4pt]}, font=\footnotesize]
  \fill[black!3, rounded corners=5pt] (-0.50,-0.55) rectangle (7.70,3.00);
  \draw[mfdline, rounded corners=5pt, line width=0.4pt] (-0.50,-0.55) rectangle (7.70,3.00);
  \node[mfdline!50!black, font=\scriptsize, anchor=east] at (7.05,0.28) {$\M$};
  \foreach \x/\y in {0/-0.15, 1.95/1.00, 3.90/0.30, 5.55/0.95, 7.20/2.78}
    {\draw[black!22, dotted, line width=0.4pt] (\x,\y) -- (\x,-1.05);}
  \draw[black!45, dashed, line width=0.5pt] (0,-0.15) arc[start angle=123.944, end angle=100.343, radius=19.005];
  \draw[->, black!55, line width=0.6pt, shorten >=3pt] (1.95,1.00) arc[start angle=230.506, end angle=270, radius=3.06607];
  \draw[->, black!55, line width=0.6pt, shorten >=3.5pt] (3.90,0.30) arc[start angle=270, end angle=313.003, radius=2.41923];
  \draw[->, black!50, line width=0.5pt, dash pattern=on 1.8pt off 1.4pt, shorten >=3.5pt] (1.95,1.00) -- (5.55,0.95)
    node[pos=0.42, sloped, below, font=\scriptsize, black!50, inner sep=3pt] {$(t-s)\,\hat u$};
  \draw[->, flowcol, line width=1.15pt] (1.95,1.00) -- (5.55,1.83)
    node[pos=0.78, sloped, below, font=\scriptsize, flowcol, inner sep=3pt] {$(t-s)\,u_\theta$};
  \draw[->, purduegold!75!black, line width=0.8pt] (1.95,1.00) -- (3.05,1.56);
  \node[font=\scriptsize, purduegold!55!black, anchor=south] at (2.95,1.60) {$\dot a_s$};
  \node[font=\scriptsize, purduegold!55!black, anchor=west] (lanch) at (1.45,2.00) {$\mathcal{L}_{\mathrm{anchor}}$};
  \draw[purduegold!55!black, line width=0.35pt] (lanch.south) -- (2.38,1.32);
  \draw[<->, eulercol, line width=0.7pt, shorten <=3pt, shorten >=3pt] (5.55,1.83) -- (5.55,0.95);
  \node[font=\scriptsize, eulercol, anchor=west] at (5.66,1.39) {$\mathcal{L}_{\mathrm{semi}}$};
  \node[circle, draw=black, line width=0.5pt, fill=white, inner sep=0pt, minimum size=3.4pt,
        label={[font=\scriptsize, yshift=1mm]below right:$a_0$}] at (0,-0.15) {};
  \node[circle, fill=black, inner sep=0pt, minimum size=3.4pt,
        label={[font=\scriptsize]above left:$a_s$}] at (1.95,1.00) {};
  \node[rectangle, draw=black!65, line width=0.5pt, fill=white, inner sep=0pt, minimum size=3.6pt,
        label={[font=\scriptsize, black!65, fill=black!3, inner sep=1pt, xshift=2mm]below:$\tilde a_r$}] at (3.90,0.30) {};
  \node[rectangle, draw=black!65, line width=0.5pt, fill=white, inner sep=0pt, minimum size=3.6pt,
        label={[font=\scriptsize, black!65]right:$\tilde a_t$}] at (5.55,0.95) {};
  \node[circle, fill=modecol, inner sep=0pt, minimum size=3.8pt,
        label={[font=\scriptsize, xshift=1mm]below left:$a_1$}] at (7.20,2.78) {};
  \draw[black!55, line width=0.4pt] (0,-1.05) -- (7.20,-1.05);
  \foreach \x/\l in {0/$0$, 1.95/$s$, 3.90/$r$, 5.55/$t$, 7.20/$1$}
    {\draw[black!55, line width=0.4pt] (\x,-0.98) -- (\x,-1.12); \node[below, inner sep=1.5pt, font=\scriptsize] at (\x,-1.12) {\l};}
\end{tikzpicture}
\caption{
  The Riemannian flowmap consistency objective~\eqref{eq:lsemi} and Riemannian conditional flow matching anchor~\eqref{eq:lanchor} are schematically shown.
  The horizontal axis represents time.
  A point $a_s$ is drawn on the conditional geodesic (dashed) from a prior sample $a_0$ to a demonstrated chunk $a_1$.
  The two flow maps through the intermediate time $r$ are rolled out without gradient (grey arcs), and their endpoint $\tilde a_t$ defines the chord target $\hat u$ of~\eqref{eq:target}.
  Tangent vectors at $a_s$ are denoted by straight arrows: the frozen chord $(t-s)\hat u = \Log_{a_s}(\tilde a_t)$ (\textcolor{black!50}{\textbf{dashed grey}}), the single jump $(t-s)\,u_\theta(a_s,s,t)$ (\textcolor{flowcol}{\textbf{teal}}) that the gradient reaches, and the geodesic velocity $\dot a_s$ (\textcolor{purduegold!75!black}{\textbf{gold}}) against which $u_\theta(a_s,s,s)$ is anchored.
  $\mathcal{L}_{\mathrm{semi}}$ penalizes the residual between the jump and the chord.
}\label{fig:semigroup}
\end{figure}
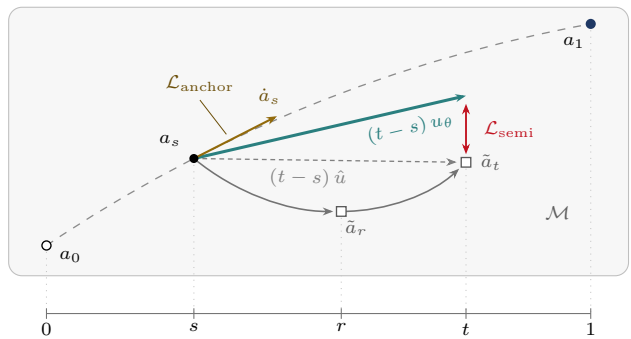

We remove that solver dependence by learning the flow map itself, so that transport over a finite interval becomes a single NFE\@.
For $0 \le s \le t \le 1$, let $\Phi_{s,t} : \M \to \M$ be the flow of the marginal field, which carries the marginal at time $s$ to the marginal at time $t$, and define the average velocity
\begin{equation}
  u(a, s, t) := \frac{1}{t - s}\Log_a\!\big(\Phi_{s,t}(a)\big) \in T_a\M ,
  \label{eq:avgvel}
\end{equation}
the geodesic chord from $a$ to its image rescaled by the interval length~\citep{Woo2026RiemannianMeanFlow}, which is well defined whenever $\Phi_{s,t}(a)$ lies within the injectivity radius of $\Exp_a$.
The map is recovered from~\eqref{eq:avgvel} by $\Phi_{s,t}(a) = \Exp_a\!\big((t-s)\, u(a,s,t)\big)$, and the diagonal limit $u(a,s,s) = v(a,s)$ returns the instantaneous field; hence~\eqref{eq:avgvel} extends rather than replaces the field that RFMP learns.
Because $\Phi_{s,t}$ is the evolution operator of a time-dependent field rather than a one-parameter group, it satisfies the two-parameter composition law $\Phi_{r,t} \circ \Phi_{s,r} = \Phi_{s,t}$ for every $s \le r \le t$, which in terms of the average velocity is the semigroup identity
\begin{equation}
  u(a, s, t) = \frac{1}{t - s}\Log_a\!\Big( \Phi_{r,t}\big(\Phi_{s,r}(a)\big) \Big) .
  \label{eq:semigroup}
\end{equation}
Identity~\eqref{eq:semigroup} is algebraic, as it contains no differential terms, which separates it from the differential MeanFlow identity and its Riemannian counterpart~\citep{Geng2025MeanFlow, Zhong2026RMFManifolds}.
We train a network $u_\theta(a, s, t \mid o)$ to satisfy~\eqref{eq:semigroup}.
Given a point $a_s$ on a conditional geodesic and times $s \le r \le t$, we define the bootstrap target,
\begin{equation}
  \hat u = \frac{1}{t - s}\Log_{a_s}\!\big(\sg[\tilde a_t]\big) ,
  \label{eq:target}
\end{equation}
where $\tilde a_t$ is the composition of two flow map evaluations,
\begin{align}
  \tilde a_r &= \Exp_{a_s}\!\big((r-s)\, u_\theta(a_s, s, r \mid o)\big), \label{eq:leg1}\\
  \tilde a_t &= \Exp_{\tilde a_r}\!\big((t-r)\, u_\theta(\tilde a_r, r, t \mid o)\big) . \label{eq:leg2}
\end{align}
The composition is then regressed via a consistency term:
\begin{equation}
  \mathcal{L}_{\mathrm{semi}}(\theta) = \E \left[ \big\| w(r) \big( u_\theta(a_s, s, t \mid o) - \hat u \big) \big\|_{a_s}^2 \right],
  \label{eq:lsemi}
\end{equation}
where $\sg$ denotes the stop-gradient operator.
The construction is illustrated in Fig.~\ref{fig:semigroup}.

Identity~\eqref{eq:semigroup} holds for any flow and is therefore data-free; in particular $u \equiv 0$ satisfies it, and hence~\eqref{eq:lsemi} cannot by itself determine the field.
A Riemannian Conditional Flow Matching anchor supplies the data boundary at $t=s$,
\begin{equation}
  \mathcal{L}_{\mathrm{anchor}}(\theta) = \E \left[ \big\| w(s) \big( u_\theta(a_s, s, s \mid o) - \dot a_s \big) \big\|_{a_s}^2 \right].
  \label{eq:lanchor}
\end{equation}
The weight $w(\tau) = (1-\tau) / \max(1-\tau, \epsilon)$ with $\epsilon = 0.1$ multiplies the residual inside the norm, and hence enters the squared loss as $w^2$~\citep{Woo2026RiemannianMeanFlow}.

\subsubsection{Velocity- and Endpoint-Prediction Parameterizations}\label{sssec:parameterization}
The network cannot emit $u_\theta$ directly, since it is a tangent vector at $a$.
We therefore parameterize the network head $h_\theta(a, s, t \mid o)$ and project its output onto the appropriate subspace.
\emph{Endpoint prediction} reads the head as a point of $\M$ and sets
\begin{equation}
  u_\theta(a, s, t \mid o) = \frac{1}{1 - s}\Log_a\!\big(\Pi_{\M}\, h_\theta\big) ,
  \label{eq:x1}
\end{equation}
where $\Pi_{\M}$ is the closest-point projection onto $\M$, a normalization on the $\Sph{3}$ factor and the identity on the Euclidean ones.
\emph{Velocity prediction} reads the head as an element of $T_a\M$ and sets $u_\theta(a, s, t \mid o) = \Pi_{T_a\M}\, h_\theta$, the orthogonal projection onto the tangent space.
The two differ in where the manifold constraint is imposed, on $\M$ or on $T_a\M$, and in the denominator they carry: the endpoint head of~\eqref{eq:x1} divides by $1-s$ and therefore requires the weight $w$, whereas velocity prediction has no denominator and does not.

Endpoint prediction is literal only at $t = 1$.
For $t < 1$ the target of~\eqref{eq:lsemi} is $\Log_{a_s}(\tilde a_t)/(t-s)$, and the optimal head is therefore $\Exp_{a_s}\!\big(\tfrac{1-s}{t-s} \Log_{a_s}(\tilde a_t)\big)$, the $s \to t$ geodesic continued at constant speed to time $1$.
On a curved manifold this agrees with $\Phi_{s,1}(a_s)$ only when the marginal flow is itself geodesic on $[s,1]$; hence away from $t = 1$ the head is an endpoint estimate used to define the reparameterized velocity.
We report how both parameterizations affect performance across tasks and environments.

\subsubsection{Overall Training Objective}\label{sssec:objective}
The training objective is the weighted sum
\begin{equation}
  \mathcal{L}(\theta) = \lambda_{\mathrm{anchor}}\, \mathcal{L}_{\mathrm{anchor}}(\theta) + \lambda_{\mathrm{semi}}\, \mathcal{L}_{\mathrm{semi}}(\theta) ,
  \label{eq:total}
\end{equation}
with $\lambda_{\mathrm{anchor}} > 0$ required, since~\eqref{eq:lanchor} is the only term that observes data.
Algorithm~\ref{alg:training} states one gradient step.
It draws one anchor time and one ordered triple, and costs four NFEs: the anchor, the single jump that carries the gradient, and the two frozen flow map evaluations, of which only the first two build a graph.

\begin{algorithm}[t]
\caption{Training step for Riemannian MeanFlow Policy}\label{alg:training}
\begin{algorithmic}[1]
  \Require manifold $\M$, prior $\rho_0$, demonstrations $\rho_1$, weights $\lambda_{\mathrm{anchor}}, \lambda_{\mathrm{semi}}$
  \State sample $(o, a_1) \sim \rho_1$, $a_0 \sim \rho_0$
  \State encode observation $c \gets \mathrm{enc}_\theta(o)$
  \LineComment{anchor term}
  \State sample $s_a$; set $a_{s_a}, \dot a_{s_a} \gets \gamma(s_a), \dot\gamma(s_a)$
  \State $\mathcal{L}_{\mathrm{anchor}} \gets \| w(s_a)\,( u_\theta(a_{s_a}, s_a, s_a \mid c) - \dot a_{s_a}) \|^2_{a_{s_a}}$
  \LineComment{semigroup term}
  \State sample $s \le r \le t$; set $a_s \gets \gamma(s)$
  \State $\tilde a_r \gets \Exp_{a_s}((r-s) u_\theta(a_s, s, r \mid c))$ \Comment{no grad}
  \State $\tilde a_t \gets \Exp_{\tilde a_r}((t-r) u_\theta(\tilde a_r, r, t \mid c))$ \Comment{no grad}
  \State $\mathcal{L}_{\mathrm{semi}} \gets \| w(r)\,( u_\theta(a_s, s, t \mid c) - \Log_{a_s}(\tilde a_t)/(t-s)) \|^2_{a_s}$
  \State update $\theta$ on $\lambda_{\mathrm{anchor}} \mathcal{L}_{\mathrm{anchor}} + \lambda_{\mathrm{semi}} \mathcal{L}_{\mathrm{semi}}$
\end{algorithmic}
\end{algorithm}

\subsection{Implementation Details}\label{ssec:implementation}

For our method and the baselines, we use the one-dimensional conditional UNet of Diffusion Policy over the action-chunk axis, with feature-wise linear modulation on the encoded observation~\citep{Chi2025DiffusionPolicy, Ding2025RFMP}.
State-based settings pass the observation stack directly; vision-based settings encode each camera with a ResNet-18 into a 512-dimensional feature.
RFMP and the diffusion baselines condition on one time and RMFP on two.
Each time input is mapped via a sinusoidal embedding and an MLP before being stacked with the conditioning.
The rest of the architecture is identical across methods.

The action manifold is $\Sph{2}$ on spherical LASA and spherical Push-T, Euclidean on the Robomimic tasks, and $\R^3 \times \Sph{3} \times \R^2$ on Franka Kitchen.
Unless stated otherwise, the prior is a standard Gaussian on Euclidean factors and a wrapped Gaussian on spherical factors, centered at the Fr\'echet mean of the demonstrated points of that factor.
On Franka Kitchen that center also pins the quaternion sign: unit quaternions double cover $\SO{3}$ and forward kinematics returns $q$ or $-q$ indifferently, which would otherwise split the demonstrations across the two hemispheres of $\Sph{3}$.

The anchor time is drawn from $\mathrm{Beta}(1.9, 1)$, mixed with a uniform draw at probability $0.02$ and rescaled to $[0.01, 1]$, which concentrates the data term near the end of the path, where the conditional endpoint is most informative about the head; only the last $\epsilon$ of that range is damped by the pole weight.
For the semigroup term the interval length is drawn from $\mathrm{Beta}(1.5, 1.5)$ on $[0.01, 1]$ and placed uniformly, with $r$ at its midpoint; $s$ and $r$ are then contracted by $0.99$ while $t$ is left untouched, and the target time therefore reaches $1$ as inference requires while the frozen rollout is never evaluated at the degenerate $r = 1$.
The pole weight uses $\epsilon = 0.1$, and both loss weights in~\eqref{eq:total} are $1$, except on spherical LASA, where $\lambda_{\mathrm{semi}} = 5$.
We train with AdamW under a 500-step linear warmup followed by a cosine decay, and evaluate an exponential moving average of the weights.
Batch sizes and epoch budgets follow~\citet{Ding2025RFMP}, and are identical across the methods compared within a benchmark.

\begin{figure*}[!t]
\centering\vspace{2mm}
\begin{subfigure}[b]{0.2113\textwidth}
  \centering
  \includegraphics[width=\linewidth]{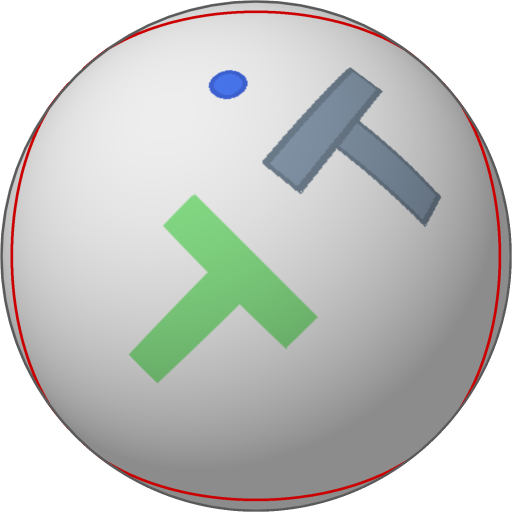}
  \caption{Spherical Push-T}\label{fig:env_pusht}
\end{subfigure}
\hfill
\begin{subfigure}[b]{0.2113\textwidth}
  \centering
  \includegraphics[width=\linewidth]{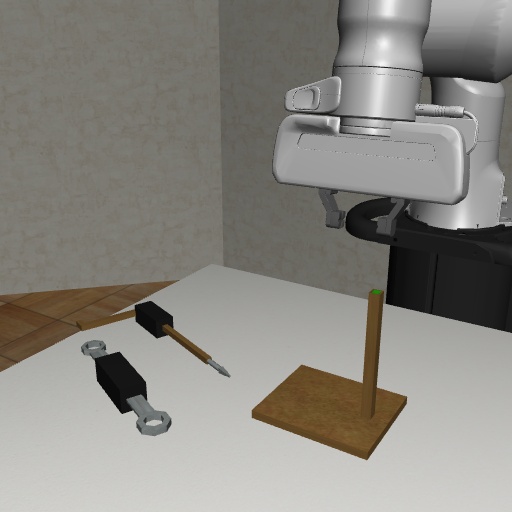}
  \caption{Robomimic Tool Hang}\label{fig:env_toolhang}
\end{subfigure}
\hfill
\begin{subfigure}[b]{0.2113\textwidth}
  \centering
  \includegraphics[width=\linewidth]{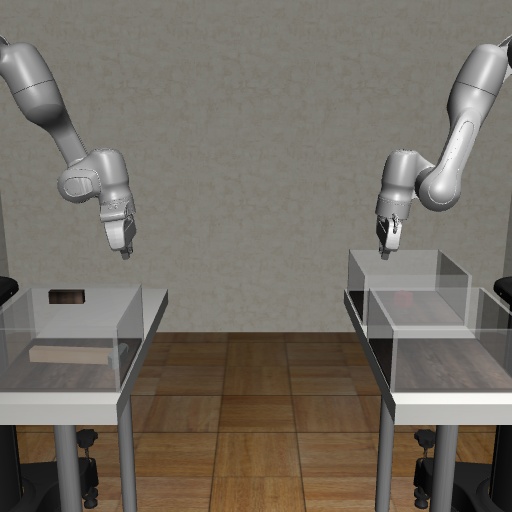}
  \caption{Robomimic Transport}\label{fig:env_transport}
\end{subfigure}
\hfill
\begin{subfigure}[b]{0.3163\textwidth}
  \centering
  \includegraphics[width=\linewidth]{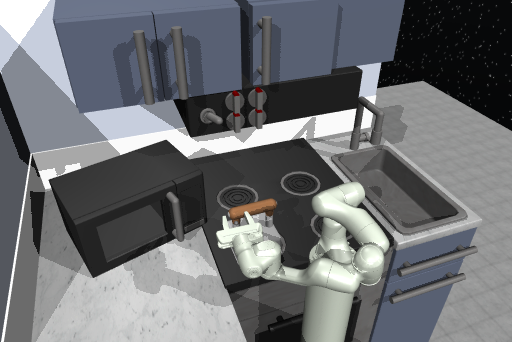}
  \caption{Franka Kitchen}\label{fig:env_kitchen}
\end{subfigure}
\caption{Simulated environments of the evaluation benchmarks.
(a) Spherical Push-T: the Push-T canvas under the stereographic map onto $\Sph{2}$, with the agent (blue), the block (grey-blue) and the goal pose (green).
(b) Robomimic Tool Hang: a single arm inserts a hook into a base and hangs a wrench on it, from a state observation.
(c) Robomimic Transport: two arms transfer a hammer from a covered container to a target bin on the opposite shelf, from camera observations.
(d) Franka Kitchen: the policy commands an end-effector pose on $\R^3 \times \Sph{3} \times \R^2$, and an episode counts as a success when at least four of the seven subtasks are completed.
}\label{fig:envs}
\end{figure*}

\section{Experiments}\label{sec:experiments}

We ask whether a learned action flow map on a Riemannian manifold retains task performance at a limited sampling budget under different settings.
We evaluate the proposed formulation against baselines for Riemannian generative policy learning on four simulated benchmarks and one trajectory-reproduction task.

\subsection{Experimental Setup}\label{ssec:exp_setup}

\subsubsection{Tasks and Datasets}\label{sssec:tasks}
We use four simulated benchmarks (cf. Fig.~\ref{fig:envs}) and one trajectory-reproduction task (cf. Fig.~\ref{fig:lasa}) that separate the effect of the training objective from the effect of the geometry of the action space.
Spherical LASA is the trajectory-reproduction benchmark of~\citet{Braun2024RFMP} wherein the LASA handwriting demonstrations~\citep{KhansariZadeh2011SEDS} are projected onto $\Sph{2}$.
We follow that protocol with $M = 7$ demonstrations of $T_m = 200$ timesteps per shape, including the multimodal shape, and report the dynamic time warping distance (DTW) to the demonstrations and the jerkiness (Jerk) of the generated trajectory.
Spherical Push-T replaces the planar agent target of Push-T by its stereographic projection onto $\Sph{2}$ and retains the planar coverage reward of the Euclidean environment~\citep{Chi2025DiffusionPolicy, Ding2025RFMP}; we report the coverage score of the target region.
From the Robomimic suite~\citep{Mandlekar2021Robomimic}, we use the Tool Hang and Transport environments.
Tool Hang is a state-based environment with a 7-DoF single-arm action space, and Transport comprises both state and vision-based observations with a 14-DoF bimanual action space.
Franka Kitchen~\citep{Gupta2019RelayPolicy} benchmarks long-horizon, multi-step planning where error in a single action chunk propagates into every subtask that follows.
Here, we use the product-manifold setting of~\citet{Ding2025RFMP}, where the policy commands an end-effector pose on $\R^3 \times \Sph{3} \times \R^2$.
A success corresponds to an episode in which at least four of the seven subtasks are completed.

The benchmarks form two regimes.
On spherical LASA, spherical Push-T, and Franka Kitchen every method is manifold-native, which tests the learning of manifold-valued action sequences.
On the Robomimic tasks, the action space is Euclidean and the Riemannian policies therefore reduce to their Euclidean counterparts, which tests the generality of the Riemannian formulations.

\subsubsection{Baselines}\label{sssec:baselines}
We benchmark our method's performance against the following baselines:
\begin{itemize}
  \item \textbf{Riemannian Flow Matching Policy} (RFMP)~\citep{Braun2024RFMP, Ding2025RFMP}:
  It regresses the instantaneous velocity field of a geodesic probability path on the action manifold and generates via the exponential-map Euler scheme of Section~\ref{sssec:rfmp}.
  It shares that manifold construction with RMFP and differs only in learning the velocity field instead of the flow map, which isolates the effect of the objective.
  This is the primary baseline we compare against.
  \item \textbf{Riemannian Diffusion Policy} (RDP): our own implementation, since no manifold-native diffusion policy exists to the best of our knowledge.
  It adapts Riemannian score-based generative modeling~\citep{DeBortoli2022RiemannianScore, Huang2022RiemannianDiffusion} to visuomotor policy learning, replacing the Euclidean forward noising of Diffusion Policy by a forward process on $\M$ and a tangent-space denoiser, after the construction used for $\SE{3}$ generation by~\citet{Yim2023SE3Diffusion,Urain2023SE3DiffusionFields}.
  It supplies the diffusion comparison on manifold-constrained learning benchmarks.
  \item \textbf{Diffusion Policy} (DP)~\citep{Chi2025DiffusionPolicy}: the Euclidean formulation.
  It generates action chunks by denoising on an unconstrained action space under receding-horizon control.
  It is the diffusion baseline on the Robomimic tasks, where the reduction of Section~\ref{sssec:tasks} makes RDP coincide with it under a DDIM sampler~\citep{Song2021DDIM}.
\end{itemize}

\subsection{Performance Evaluation}\label{ssec:exp_eval}

\begin{table}[t]
  \centering\vspace{2mm}\footnotesize
  \caption{LASA on $\Sph{2}$, ten characters.
  Median DTW to the demonstration, and the executed path's jerk (the demonstration
  equals $1.0$); lower is better. ($\lambda_{\mathrm{semi}} = 5$)
  }\label{tab:lasa}
  \setlength{\tabcolsep}{2.5pt}
  \begin{tabular}{lcccccccc}
    \toprule
    \multirow{2}{*}[\cmidrowshift]{Method} & \multicolumn{2}{c}{NFE 1} & \multicolumn{2}{c}{NFE 2} & \multicolumn{2}{c}{NFE 5} & \multicolumn{2}{c}{NFE 10} \\
    \cmidrule(lr){2-3} \cmidrule(lr){4-5} \cmidrule(lr){6-7} \cmidrule(lr){8-9}
    & DTW & Jerk & DTW & Jerk & DTW & Jerk & DTW & Jerk \\
    \midrule
    RFM       & 0.090 & 857 & \underline{0.020} & 153 & \textbf{0.016} & \textbf{83} & \textbf{0.016} & \textbf{59} \\
    \addlinespace
    RMF-$v$   & \underline{0.033} & \underline{317} & \textbf{0.017} & \textbf{96} & \underline{0.019} & \underline{86} & \underline{0.020} & \underline{82} \\
    RMF-$x_1$ & \textbf{0.022} & \textbf{153} & 0.022 & \underline{146} & 0.026 & 143 & 0.024 & 113 \\
    \bottomrule
  \end{tabular}
\end{table}

We evaluate our method against the baselines across budgets of $1$, $2$, $5$ and $10$ NFEs per action chunk, reported in Tables~\ref{tab:lasa}--\ref{tab:budget_transport}.
In the tables, RFM abbreviates RFMP and RMF abbreviates the proposed RMFP, whose velocity- and endpoint-prediction parameterizations of Section~\ref{sssec:parameterization} appear as RMF-$v$ and RMF-$x_1$; RDP-$\epsilon$ and RDP-$x$ are the variants of RDP under its noise- and sample-prediction heads.
On the Franka Kitchen task, RMF and RFM are additionally trained and evaluated under two different priors on the $\Sph{3}$ factor: the suffix -uni draws that factor uniformly on the sphere, and -hemi uniformly on the closed hemisphere that the sign convention of Section~\ref{ssec:implementation} selects.
A score-based model draws from the terminal law of its forward process rather than from a prior that can be chosen, and the RDP entries therefore carry no prior suffix.

\subsubsection{LASA Handwriting Dataset}\label{sssec:res_lasa}

Each run is trained for $600$ epochs and uses a uniform prior over the hemisphere.
In the few-step regime ($\mathrm{NFE}\in\{1,2\}$), both RMF variants produce smoother trajectories than RFM across the ten tested characters (Table~\ref{tab:lasa}), and at $\mathrm{NFE}=1$ both also attain a lower median DTW, $0.022$ for RMF-$x_1$ and $0.033$ for RMF-$v$ against $0.090$ for RFM.
Fig.~\ref{fig:lasa} shows this difference: RMF traces a smoother trajectory and follows the demonstrations more closely than RFM.
At $\mathrm{NFE}\geq5$ the ordering reverses: RFM attains the lowest median DTW, $0.016$ at both budgets against $0.019$ and $0.020$ for RMF-$v$, at the cost of the additional network evaluations.

\begin{table}[!t]
\centering\vspace{2mm}
\caption{Spherical Push-T coverage score against the sampling budget}\label{tab:budget_pusht}
\footnotesize
\begin{tabular}{@{}lcccc@{}}
\toprule
\multirow{2}{*}[\cmidrowshift]{Method} & \multicolumn{4}{c}{Coverage Score $(\%)$ $\uparrow$} \\
\cmidrule(lr){2-5}
 & $1$ NFE & $2$ NFE & $5$ NFE & $10$ NFE \\
\midrule
RDP-$\epsilon$ & 15.0 & 51.8 & 83.9 & 81.6 \\
RDP-$x$        & 73.1 & 74.0 & 76.0 & 74.4 \\
RFM            & 67.9 & 77.2 & \underline{84.7} & 83.9 \\
\addlinespace
RMF-$v$        & \textbf{78.5} & \textbf{81.0} & \textbf{85.7} & \underline{84.1} \\
RMF-$x_1$      & \underline{76.0} & \underline{78.3} & 81.7 & \textbf{85.0} \\
\bottomrule
\end{tabular}
\end{table}

\begin{table}[!t]
\centering\vspace{2mm}
\caption{Franka Kitchen success rate against the sampling budget}\label{tab:budget_kitchen}
\footnotesize
\begin{tabular}{@{}lcccc@{}}
\toprule
\multirow{2}{*}[\cmidrowshift]{Method} & \multicolumn{4}{c}{Success Rate $(\%)$ $\uparrow$} \\
\cmidrule(lr){2-5}
 & $1$ NFE & $2$ NFE & $5$ NFE & $10$ NFE \\
\midrule
RDP-$\epsilon$ & 3.7 & 19.7 & 36.9 & 47.1 \\
RDP-$x$        & 23.7 & 41.7 & 43.1 & 39.1 \\
RFM-uni        & 14.9 & 40.3 & 44.3 & 41.1 \\
RFM-hemi       & 20.6 & 41.7 & \underline{47.1} & \underline{48.6} \\
\addlinespace
RMF-$v$-uni    & 34.9 & 41.7 & 44.3 & 45.1 \\
RMF-$v$-hemi   & 40.6 & 41.1 & 46.0 & 43.4 \\
RMF-$x_1$-uni  & \underline{41.4} & \underline{43.4} & 44.6 & \textbf{48.9} \\
RMF-$x_1$-hemi & \textbf{41.7} & \textbf{46.0} & \textbf{47.7} & 46.9 \\
\bottomrule
\end{tabular}
\end{table}

\subsubsection{Spherical Push-T}\label{sssec:res_pusht}
At $\mathrm{NFE}=1$, both RMF variants exceed every baseline (Table~\ref{tab:budget_pusht}): RMF-$v$ reaches $78.5$ against $67.9$ for RFM and $73.1$ for RDP-$x$.
The margin closes as the budget grows, from $3.8$ points at $\mathrm{NFE}=2$ to $1.0$ at $\mathrm{NFE}=5$, and at $\mathrm{NFE}=10$ the leading entry of the two RMF variants is $85.0$ against $83.9$ for the baselines.
The two diffusion variants fail differently.
RDP-$\epsilon$ collapses to $15.0$ at $\mathrm{NFE}=1$ and closes on the other methods only at $\mathrm{NFE}=5$, whereas RDP-$x$, whose head emits a point of the action space rather than the perturbation applied to it, does not collapse; it spans $2.9$ points across the whole sweep ($73.1$ to $76.0$) and remains below the $\mathrm{NFE}=10$ entry of every other method.

\subsubsection{Franka Kitchen}\label{sssec:res_kitchen}
At $\mathrm{NFE}=1$, the weakest RMF variant, RMF-$v$-uni at $34.9$, exceeds the strongest baseline, RDP-$x$ at $23.7$, by $11.2$ points (Table~\ref{tab:budget_kitchen}).
Against the prior-matched flow-matching baseline the margin is $21.1$ points for RMF-$x_1$-hemi over RFM-hemi and $26.5$ for RMF-$x_1$-uni over RFM-uni.
The margin again closes with the budget, from $4.3$ points at $\mathrm{NFE}=2$ to $0.6$ at $\mathrm{NFE}=5$, and at $\mathrm{NFE}=10$ the leading entries are $48.9$ against $48.6$.
No entry exceeds $48.9$: the long-horizon, multi-step reasoning requirement of Section~\ref{sssec:tasks} holds every method far from a perfect score, and the $\mathrm{NFE}=1$ margins are correspondingly wider than on Push-T.
RFM-hemi loses $28.0$ points at $\mathrm{NFE}=1$ against its own best budget, and RDP-$\epsilon$ collapses to $3.7$ and closes only at $\mathrm{NFE}=10$; RDP-$x$ is stable from $\mathrm{NFE}=2$ upward, where it spans $4.0$ points, and loses $18.0$ points at one against its $\mathrm{NFE}=2$ entry.
The hemisphere prior is worth most at $\mathrm{NFE}=1$ and to the velocity head: $5.7$ points to RFM ($20.6$ against $14.9$) and $5.7$ to RMF-$v$ ($40.6$ against $34.9$), against $0.3$ to RMF-$x_1$ ($41.7$ against $41.4$).
The endpoint prediction variant RMF-$x_1$ regresses, at the prior sample, a point of $\M$ rather than a tangent vector, which is consistent with it being the variant least sensitive to the prior at the budget where a single transport must cover the whole interval.

\begin{table}[!t]
\centering\vspace{2mm}
\caption{Robomimic Tool Hang success rate against the sampling budget}\label{tab:budget_toolhang}
\footnotesize
\begin{tabular}{@{}lcccc@{}}
\toprule
\multirow{2}{*}[\cmidrowshift]{Method} & \multicolumn{4}{c}{Success Rate $(\%)$ $\uparrow$} \\
\cmidrule(lr){2-5}
 & $1$ NFE & $2$ NFE & $5$ NFE & $10$ NFE \\
\midrule
DP        & 0.0 & 46.0 & \underline{64.0} & \textbf{74.0} \\
RFM       & \underline{60.0} & \underline{60.0} & 60.0 & 64.0 \\
\addlinespace
RMF-$v$   & \textbf{64.0} & \textbf{76.0} & \textbf{72.0} & \textbf{74.0} \\
RMF-$x_1$ & 48.0 & 52.0 & 60.0 & \textbf{74.0} \\
\bottomrule
\end{tabular}
\end{table}

\begin{table}[!t]
\centering\vspace{2mm}
\caption{Robomimic Transport success rate against the sampling budget}\label{tab:budget_transport}
\footnotesize
\begin{tabular}{@{}lcccc@{}}
\toprule
\multirow{2}{*}[\cmidrowshift]{Method} & \multicolumn{4}{c}{Success Rate $(\%)$ $\uparrow$} \\
\cmidrule(lr){2-5}
 & $1$ NFE & $2$ NFE & $5$ NFE & $10$ NFE \\
\midrule
DP        & 0.0 & 82.0 & \textbf{96.0} & 82.0 \\
RFM       & \textbf{94.0} & 86.0 & 88.0 & 86.0 \\
\addlinespace
RMF-$v$   & \underline{90.0} & \textbf{92.0} & \underline{92.0} & \textbf{92.0} \\
RMF-$x_1$ & \textbf{94.0} & \underline{90.0} & \underline{92.0} & \underline{90.0} \\
\bottomrule
\end{tabular}
\end{table}

\subsubsection{Robomimic Tool Hang}\label{sssec:res_toolhang}
The Robomimic suite has a Euclidean action space, which removes the geometric construction from the comparison.
RMF-$v$ leads at every budget $\leq 5$ NFE (Table~\ref{tab:budget_toolhang}), by $4.0$ points over RFM at $\mathrm{NFE}=1$, by $16.0$ at $\mathrm{NFE}=2$ and by $8.0$ over DP at $\mathrm{NFE}=5$, and its two-evaluation entry of $76.0$ is the largest in the table.
The $\mathrm{NFE}=1$ entry of RMF-$v$ equals the $\mathrm{NFE}=5$ entry of DP and the $\mathrm{NFE}=10$ entry of RFM, both at $64.0$.
DP collapses to $0.0$ at $\mathrm{NFE}=1$ and recovers monotonically, reaching $74.0$ only at $\mathrm{NFE}=10$, whereas RFM holds $60.0$ through $\mathrm{NFE}=5$ and reaches $64.0$ at $\mathrm{NFE}=10$, the only method in the table that does not attain the $74.0$ the other three reach.
On a Euclidean action space a one-step RFM sample is the conditional mean of the demonstrations, and the task metric nonetheless registers a loss of only $4.0$ points against its own $\mathrm{NFE}=10$ entry.

\subsubsection{Robomimic Transport}\label{sssec:res_transport}
The Transport task from the Robomimic suite carries the Euclidean reduction into the other observation modality and onto a 14-DoF bimanual action space conditioned on proprioception and multi-view camera observations.
The benchmark does not separate the three flow-based arms (Table~\ref{tab:budget_transport}), whose entries lie between $86.0$ and $94.0$ with no pairwise difference at a fixed budget above $6.0$ points; it separates the diffusion baseline at $\mathrm{NFE}=1$ alone, where DP is $0.0$ and no other arm falls below $90.0$.
RMF-$v$ is the one arm that is constant for $\mathrm{NFE}\geq2$, at $92.0$.
Here the task metric registers no loss: the one-evaluation entry of $94.0$ is the largest RFM attains.
Hence, alongwith the Tool Hang benchmark, we observe that the semigroup objective trains unmodified and the low-budget margin it holds on the unsaturated member of the pair is therefore not an artifact of curvature.

Across the benchmarks, we note that the flow map consistency objective buys the low-budget regime rather than a higher ceiling: on spherical Push-T and Franka Kitchen its margin over flow matching is largest at $\mathrm{NFE}=1$ and falls to at most $1.0$ point at $\mathrm{NFE}=5$.
Furthermore, it flattens the response to varying NFE budgets.
Read as a range over the four budgets of those two benchmarks, each RMF arm spans at most $10.2$ points, against $16.8$ to $29.4$ for the flow-matching arms and $43.4$ to $68.9$ for RDP-$\epsilon$.
The RMF entries at 2, 5 and 10 NFE query interval lengths inside the dense part of the training law of Section~\ref{ssec:implementation}, and their agreement with the $\mathrm{NFE}=1$ entry, within $7.2$ points for RMF-$v$ on Push-T and $6.0$ for RMF-$x_1$-hemi on Franka Kitchen, is what evidences that the extrapolation from those lengths to the full interval holds.
The two parameterizations do not order consistently.
At $\mathrm{NFE}=1$, RMF-$v$ leads RMF-$x_1$ by $2.5$ points on spherical Push-T and by $16.0$ points on Tool Hang, whereas RMF-$x_1$ leads by $6.5$ points under the uniform prior and by $1.1$ points under the hemisphere prior on Franka Kitchen.
The ordering tracks the dimension of the action rather than its curvature: the endpoint parameterization leads on the nine-coordinate product action of Franka Kitchen and trails on the seven-dimensional Tool Hang action and on spherical Push-T.
This matches the direction reported by~\citet{Woo2026RiemannianMeanFlow}, who find endpoint prediction the more stable head in high dimension, albeit on a quantity this protocol does not measure.
Transport places its 14-DoF action on the leading side, by $4.0$ points at $\mathrm{NFE}=1$, a margin that benchmark does not resolve.
Reporting both heads is therefore not redundant: as Section~\ref{sssec:parameterization} states, the two impose the manifold constraint at different points of the forward pass and have no reason to agree.

\subsection{Real-world Deployment}\label{ssec:rw}
To demonstrate the applicability of our approach to practical robotic manipulation, we evaluate the learned policies on a real-world robotic platform.
Our hardware setup consists of a 6-DoF I2RT YAM robotic arm equipped with a linear parallel-jaw gripper and controlled over CAN by an NVIDIA Jetson Thor, which also hosts policy inference.
The workspace is observed using two RGB-D cameras: an external Orbbec Femto Mega ($1280 \times 720$ at $30$~Hz) mounted in front of the workspace to provide a global third-person view, and an Intel RealSense D405 ($640 \times 480$ at $30$~Hz) mounted on the robot wrist to provide close-range observations during manipulation.
The external camera is hand-eye calibrated with respect to the robot base, and both cameras remain fixed throughout data collection and evaluation.

For real-world evaluation, we collect additional demonstrations directly on the physical system across a set of manipulation tasks.
The demonstrations contain synchronized visual observations and robot trajectories, which are used to train the policies before deployment on the same hardware platform.
The resulting policies are executed on the physical robot using observations from both camera views together with robot proprioception.
Fig.~\ref{fig:rw} shows an example of a real-world rollout of RMF.
We provide video demonstrations in the supplementary material.

\begin{figure}[t]
  \centering\vspace{2mm}
  \includegraphics[width=0.75\linewidth]{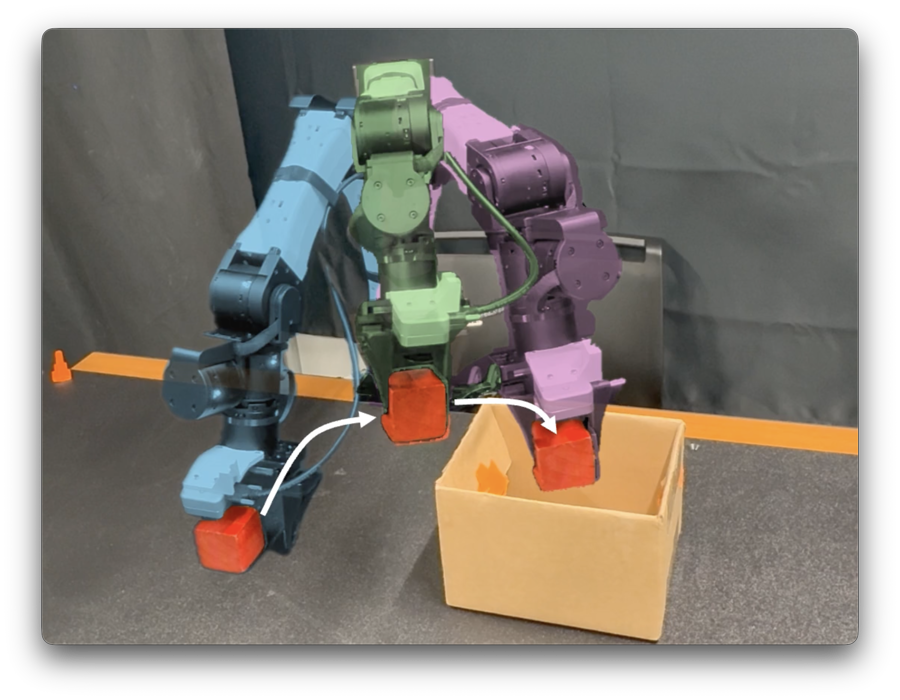}
  \caption{Riemannian MeanFlow Policy executed in a real-world manipulation task.
  The policy is trained on $60$ real-world demonstrations and deployed on the YAM robotic platform.
  Representative stages of the rollout show the robot approaching and grasping the object, transporting it to a goal state, and placing it into the target container.
  }\label{fig:rw}
\end{figure}

\section{Conclusion}\label{sec:conclusion}

This work addresses the sampling cost of geometrically constrained visuomotor policies by learning the flow map of the generative probability path instead of the velocity field that induces it.
The Riemannian MeanFlow Policy regresses the average velocity on the robot action manifold under an algebraic semigroup identity imposed as a flow map consistency, anchored to the demonstrations by Riemannian Conditional Flow Matching.
The identity is algebraic and carries no differential terms, which keeps training first-order and stable.
The formulation trains in a single stage without distilling from a teacher, and the map queried by a one-step sample is the map the objective regresses.

\section*{Acknowledgment}
Claude Code was used for language polishing and editorial assistance. All technical content, equations, experimental results, and final manuscript text were reviewed and verified by the authors, who assume full responsibility for the final manuscript.

\bibliographystyle{IEEEtranN}
\bibliography{references} %

\end{document}